\documentclass{article}

\usepackage{microtype}
\usepackage{graphicx}
\usepackage{subcaption}
\usepackage{booktabs} 

\usepackage{hyperref}
\usepackage{multirow}

\usepackage[preprint]{icml2026}

\usepackage{amsmath}
\usepackage{amssymb}
\usepackage{mathtools}
\usepackage{amsthm}
\usepackage[capitalize,noabbrev]{cleveref}

\theoremstyle{plain}

\theoremstyle{definition}

\theoremstyle{remark}

\usepackage{xcolor}
\usepackage{float}
\usepackage{xspace}
\newcommand{\eg}{\emph{e.g.}\xspace}
\newcommand{\ie}{\emph{i.e.}\xspace}

\usepackage[acronym]{glossaries}

\newacronym{mcmc}{MCMC}{Markov chain Monte Carlo}
\newacronym{vae}{VAE}{variational autoencoder}
\newacronym{cfg}{CFG}{classifier-free guidance}
\newacronym{cfm}{CFM}{conditional flow matching}
\newacronym{cnn}{CNN}{convolutional neural network}
\newacronym{ode}{ODE}{ordinary differential equation}
\newacronym{lde}{LDE}{latent directed evolution}
\newacronym{ggs}{GGS}{Gibbs sampling with Graph-based Smoothing}
\newacronym{gwg}{GWG}{Gibbs With Gradients}
\newacronym{plm}{pLM}{protein language model}
\newacronym{kl}{KL}{Kullback-Leibler}
\newacronym{method}{CHASE}{Conditional High-fitness Amino-acid Sequence Enhancement} 
\newacronym{aav}{AAV}{Adeno-Associated Virus}
\newacronym{vlgpo}{VLGPO}{Variational Latent Generative Protein Optimization}
\newacronym{gfp}{GFP}{Green Fluorescent Protein}
\newacronym{elbo}{ELBO}{Evidence Lower Bound}
\newacronym{ddpm}{DDPM}{denoising diffusion probabilistic model}
\newacronym{gpe}{GPE}{generative property enhancer}

\icmltitlerunning{Repurposing Protein Language Models for Latent Flow–Based Fitness Optimization}

\begin{document}

\twocolumn[
\icmltitle{Repurposing Protein Language Models for Latent Flow–Based\\ Fitness Optimization}

  \icmlsetsymbol{equal}{*}

  \begin{icmlauthorlist}
    \icmlauthor{Amaru Caceres Arroyo}{equal,eth}
    \icmlauthor{Lea Bogensperger}{equal,uzh}
    \icmlauthor{Ahmed Allam}{uzh}
    \icmlauthor{Michael Krauthammer}{uzh}
    \icmlauthor{Konrad Schindler}{eth}
    \icmlauthor{Dominik Narnhofer}{eth}
  \end{icmlauthorlist}

    \icmlaffiliation{uzh}{University of Zurich}
    \icmlaffiliation{eth}{ETH Zurich}

  \icmlcorrespondingauthor{Lea Bogensperger}{lea.bogensperger@uzh.ch}
  \icmlkeywords{Machine Learning, ICML}

  \vskip 0.3in
]

\printAffiliationsAndNotice{\icmlEqualContribution}

\begin{abstract}
Protein fitness optimization is challenged by a vast combinatorial landscape where high-fitness variants are extremely sparse. Many current methods either underperform or require computationally expensive gradient-based sampling. We present \glsentrytext{method}, a framework that repurposes the evolutionary knowledge of pretrained protein language models by compressing their embeddings into a compact latent space. By training a conditional flow-matching model with classifier-free guidance, we enable the direct generation of high-fitness variants without predictor-based guidance during the \glsentrytext{ode} sampling steps. \glsentrytext{method} achieves state-of-the-art performance on \glsentrytext{aav} and \glsentrytext{gfp} protein design benchmarks.
Finally, we show that bootstrapping with synthetic data can further enhance performance in data-constrained settings.
\end{abstract}


\section{Introduction}
\label{sec:introduction}

\begin{figure}[!ht]
\centering
\includegraphics[width=\linewidth] {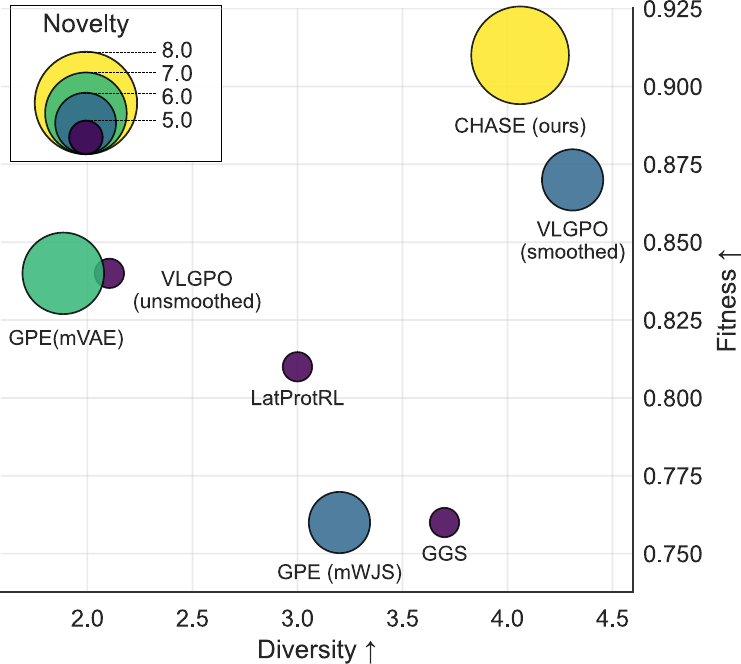} 
\caption{Fitness–diversity–novelty chart across protein optimization methods for \glsentrytext{gfp} Medium. \glsentrytext{method} achieves high fitness while maintaining diversity and novelty compared to existing baselines, whereas alternative methods tend to favor individual objectives at the expense of others.}
\label{fig:pareto}
\end{figure}

\begin{figure*}[!ht]
\centering
\includegraphics[width=\linewidth] {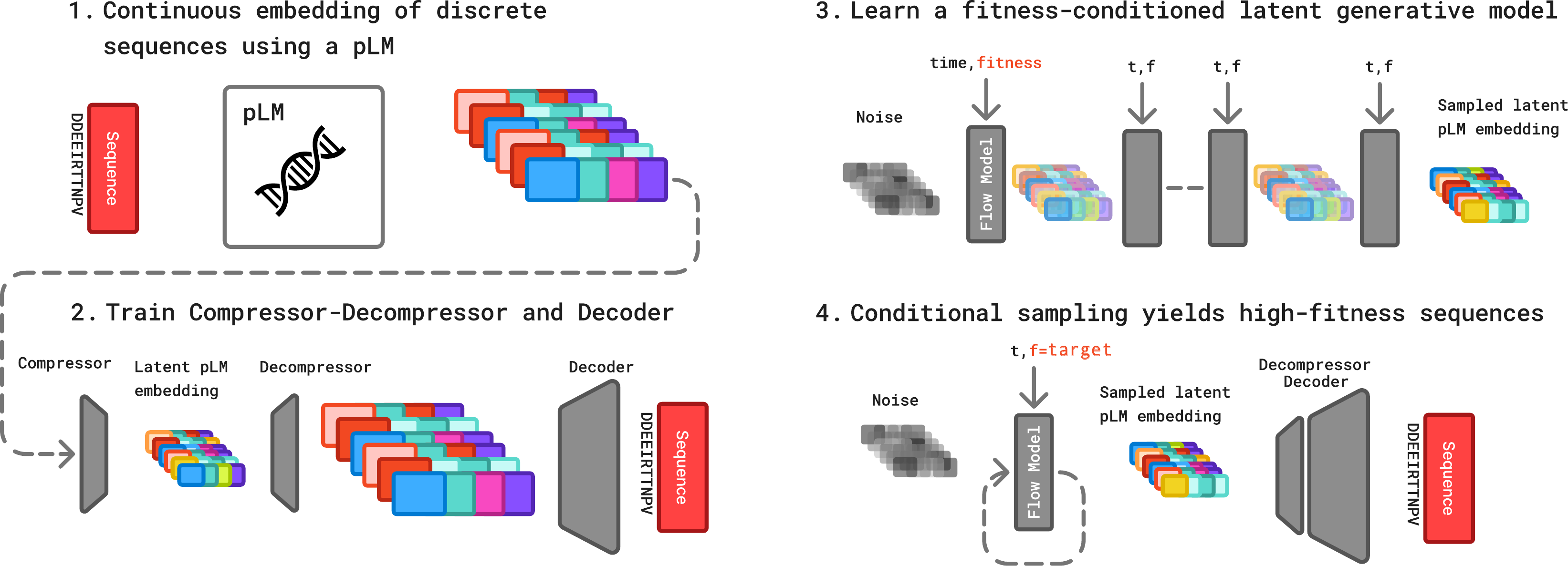} 
\caption{Overview of \glsentrytext{method}. (1) Discrete protein sequences are encoded by a pretrained \gls{plm} and (2) projected into a compact latent manifold via a compression block. (3) A conditional generative model learns the probability path between noise and these embeddings, conditioned on time $t$ and fitness $f$. (4) At inference, high-fitness embeddings are sampled and mapped back to sequence space through a decompressor and a discrete decoder. Gray blocks denote modules trained in our framework.}
\label{fig:teaser}
\end{figure*}

Proteins are essential macromolecules whose primary amino acid sequences dictate their three-dimensional structures, which in turn determine their biological functions—such as stability, binding affinity, and catalytic efficiency.
Protein fitness optimization aims to enhance these properties by navigating the fitness landscape, a vast combinatorial space of potential sequences. Because the number of possible sequences grows exponentially with length, identifying variants with meaningful biological functionality remains a significant challenge~\cite{hermes}.
Traditional directed evolution \cite{directedevolution} mimics natural selection through iterative laboratory cycles; however, despite its success, it is fundamentally limited in its ability to efficiently explore this high-dimensional and often sparse landscape.

To overcome these limitations, a variety of in-silico approaches have been proposed to computationally navigate the fitness landscape and prioritize only the most promising candidates for experimental validation. These strategies range from evolutionary greedy algorithms \cite{insilicodirectedevolution, guidedexplorationfitness, adalead} as well as gradient-based sampling methods such as \gls{gwg} \cite{oops, discretemcmc} and its smoothed variants \cite{smoothgwg}, which utilize local fitness gradients to explore discrete sequence spaces.
However, navigating these landscapes remains challenging because protein sequences are discrete and rugged due to epistatic interactions, where a mutation's effect depends on its amino acid context \cite{cleve}. 

Recent methods like \gls{vlgpo} \cite{vlgpo} address these challenges by embedding proteins into a continuous latent space. While this enables smoother exploration via flow matching \cite{flowmatching}, current approaches rely on predictor-guided trajectors during sampling, which can be computationally demanding if they have to be backpropagated through~\cite{chung2022diffusion}. 
Furthermore, these models rely on \glspl{vae} that learn latent spaces de novo, failing to leverage the rich, evolutionary priors captured by large-scale pre-trained protein language models (pLMs) \cite{esm2}.

In this work, we propose \gls{method}, a framework that repurposes the expressive embedding space of \glspl{plm} into a compact latent representation inspired by the Hourglass architecture in~\cite{protflow}, which leverages the structural priors implicitly encoded in \glspl{plm}~\cite{rao2020transformer}.
Our approach is summarized graphically in \cref{fig:teaser}. Specifically, we make the following contributions:
\begin{itemize}
    \item We demonstrate that pretrained \glspl{plm} can be repurposed into a compressed, continuous latent space capable of capturing the fine-grained fitness effects of mutations within a specific protein family.
    \item We develop a framework that adapts conditional flow matching and classifier-free guidance to learn the distribution of latent protein sequences, effectively eliminating the need for external predictor models during sampling.
    \item We achieve state-of-the-art performance on two protein fitness optimization benchmarks, generating candidates with competitive fitness while maintaining sequence diversity and novelty.
    \item We propose a data augmentation and bootstrapping strategy that expands the limited training set with synthetic sequence-fitness pairs, further enhancing the search for high-fitness variants.
\end{itemize}


\section{Related Work}
\label{sec:related_work}

\paragraph{Traditional and Surrogate-Based Optimization.}
The foundational paradigm for protein engineering is directed evolution \cite{directedevolution}, which navigates fitness landscapes through physical mutagenesis and screening. This method, which can be viewed as a random walk from a wild-type starting sequence, is inherently resource-intensive: the majority of random mutations are deleterious, and the throughput of physical assays cannot scale to the exponential complexity of the sequence-to-function mapping. 

To mitigate these costs, surrogate models are increasingly employed to guide the search more efficiently \cite{adalead,ren,lde}. For instance, Adalead \cite{adalead} utilizes a black-box predictor to inform a greedy algorithm, prioritizing mutations with the highest predicted fitness gains. Similarly, LatProtRL frames the search as a reinforcement learning problem, simulating active learning by perturbing the latent space of a large \gls{plm} to identify high-fitness actions~\cite{lee2024robust}. Recently, EvolvePro was introduced~\cite{jiang2024rapid}, a powerful active learning framework which combines the computational capacity to navigate high-dimensional search spaces with empirical feedback to ground the model’s predictions.

\paragraph{Discrete Generative Models.}
Efficient optimization requires in-silico methods capable of proposing high-fitness candidates without immediate experimental feedback. As protein sequences are discrete, many generative approaches model this representation directly. Sampling-based methods, such as the walk-jump algorithm \cite{protdesignwithdiff,protdiscowalkjump}, learn distributions over one-hot encodings using a single noise level and a Tweedie step for recovery via \gls{mcmc} sampling. This framework was later extended to incorporate gradient guidance within the noisy manifold \cite{gradguidancemanifold}.

Parallel efforts, such as \gls{gwg} \cite{oops,discretemcmc} and its variant \gls{ggs} \cite{smoothgwg}, introduce gradient-informed \gls{mcmc} for discrete sequences. These methods train a separate predictor model on limited datasets to generate guiding gradients that steer the sampling process. \gls{ggs} specifically incorporates graph-based smoothing to regularize the inherently noisy fitness landscape, providing more stable optimization trajectories. 

\paragraph{Optimization in Continuous Latent Spaces.}
As sequence length increases, the search space expands exponentially, leading recent research to explore continuous latent representations to facilitate optimization \cite{protwave}. For example, \gls{lde} \cite{lde} maps sequences to a smoothed latent space, enabling gradient ascent-based optimization guided by a learned fitness predictor. Similarly, \gls{vlgpo} \cite{vlgpo} leverages the predictors established in \cite{smoothgwg} to perform informed posterior sampling, executing gradient steps toward high-fitness regions within the continuous manifold during the generative process. More recently, \gls{gpe}~\cite{pinheiroimplicit} introduced a formulation that implicitly guides generation from low-fitness to high-fitness samples through conditional density estimation.

\paragraph{Protein Language Representations.}
The efficacy of these methods relies heavily on the quality of the underlying sequence representation. \glspl{plm}, such as ESM2 \cite{esm2}, have emerged as the standard for capturing complex evolutionary patterns and implicitly learning the structural regularities of the protein space from massive sequence databases~\cite{rao2020transformer}. Recent frameworks like ProtFlow \cite{protflow} leverage these pretrained \glspl{plm} by operating within a compressed latent embedding space, enabling efficient de novo design. While ProtFlow demonstrates the power of latent-space generation, it primarily focuses on unconditional backbone and sequence generation; our work extends this paradigm by introducing direct fitness-conditioning directly into the flow-matching objective in the compressed \gls{plm} latent space.

Building on these advancements, we propose \gls{method}, a framework that leverages the rich, continuous \gls{plm} embeddings of ESM2 \cite{esm2} to establish an expressive representation space. Inspired by the architectural insights of \cite{protflow}, we project these embeddings into a compressed manifold where fitness information is directly integrated into the flow-matching objective. By learning the latent distribution conditioned on these fitness values, \gls{method} enables fast sequence generation with targeted high-fitness while preserving diversity and novelty. Furthermore, our approach supports controllable fitness sampling and facilitates synthetic data augmentation, which provides a mechanism for efficient model bootstrapping in data-constrained regimes.


\section{Method}
\label{sec:method}

\subsection{Protein Optimization}
Let $\mathcal{V}$ denote the protein vocabulary consisting of 20 canonical amino acids and associated special tokens. 
A protein sequence of length $L$ is defined as $x = (x_1, \dots, x_L) \in \mathcal{V}^L$. We define a fitness landscape as a mapping $g: \mathcal{V}^L \to \mathbb{R}$, thereby assigning a scalar value $g(x)$ to each sequence $x$ representing its biological performance.

The goal of protein fitness optimization is to identify new sequences $x$ that maximize this property, which is achieved by introducing mutations, \ie substituting amino acids in a starting sequence with any variant in $\mathcal{V}$ to enhance the fitness value $f$. 
In this work, we focus on a constrained benchmark setting where we only have access to a limited subset of sequence-fitness pairs $\mathcal{A} \subset \mathcal{A}_{\text{GT}}$~\cite{smoothgwg}. This training set $\mathcal{A}$ is restricted to narrow fitness ranges and contains only variants that maintain a significant distance from the highest-performing sequences in the full, unknown ground-truth dataset $\mathcal{A}_{\text{GT}}$. Our modeling and data augmentation are performed exclusively on $\mathcal{A}$; the ground-truth landscape $\mathcal{A}_{\text{GT}}$ is reserved solely for final evaluation to evaluate the fitness of the generated sequences using an oracle trained on $\mathcal{A}_{\text{GT}}$.

Our framework, \gls{method}, is shown in~\cref{fig:teaser}. First, we establish a continuous representation space by mapping discrete sequences $x \in \mathcal{V}^L$ to high-dimensional \gls{plm} embeddings $h \in \mathbb{R}^{L \times D}$ (1), which are subsequently compressed into a low-dimensional latent manifold $z \in \mathbb{R}^{l \times d}$ (2). Then we train a conditional flow matching model to learn the distribution $p(z\mid f)$ within this compressed space (3). Finally, we sample $z$ given a target fitness $f$ and reconstruct the resulting sequence $x'$ (4). We explain the details of each step in the following sections.

\subsection{Latent Manifold Construction}
\label{subsec:latent_const}
The \gls{vae} framework \citep{vae,betaVAE} provides a principled approach to mapping discrete protein sequences into a continuous latent space. A \gls{vae} learns a probabilistic mapping between data $x$ and a lower‐dimensional latent representation $z$ by maximizing the \gls{elbo}:
\begin{equation}
    \mathcal{L}(x;\theta) = \mathbb{E}_{q_\theta(z \mid x)}[\log p_\theta(x|z)] - \beta\, \mathrm{KL}\!\left(q_\theta(z|x) \,\|\, p(z)\right).
\end{equation} 
This objective balances reconstruction accuracy with latent space regularization, since the \gls{kl} divergence encourages the approximate posterior $q_\theta(z|x)$ to remain close to a prior $p(z)=\mathcal{N}(0, I)$.  

In \gls{method}, we implement this framework as a two-stage process (steps (1) and (2) in \cref{fig:teaser}). First, we repurpose a pretrained ESM2-8M encoder $\mathcal{E}$ \cite{esm2} without further finetuning to map the discrete input $x$ into a continuous embedding $h \in \mathbb{R}^{L \times D}$. The corresponding decoder $\mathcal{D}$ is trained via cross-entropy loss to reconstruct sequence logits from $h$.

Second, as even the smallest ESM2 embeddings ($D=320$) are high-dimensional and prone to activation sparsity \cite{geometryoflargetransformers}, we employ an additional compressor $\mathcal{C}$ and decompressor $\mathcal{R}$ pair adopted from \cite{protflow}. This architecture reduces both the feature dimension to $d \ll D$ and the sequence length to $l < L$ (\eg, $l \in \{16, 32\}$). The decompressor processes embeddings via transformer layers and down-samples the sequence length using 1D convolutions; two parallel linear layers then output the latent mean and variance to enable the $\beta$-\gls{vae} objective. The decompressor $\mathcal{R}$ symmetrically reverses this mapping from $z$ back to the reconstructed embedding space $h'$ for final decoding by $\mathcal{D}$. Detailed architectural specifications are provided in \cref{fig:detailed_architecture} (Appendix).
With the ESM2 encoder and the decoder frozen, we optimize these components using a multi-task objective:
\begin{equation}
\begin{split}
    \mathcal{L}_{\text{VAE}}= \mathcal{L}_{\text{MSE}}(h,h') + \mathcal{L}_{\text{CE}}(x,x') \\ +  \beta\cdot \text{KL}(q_{\mu}(z|x)||p(z)), 
\end{split}
\label{eq:comb_obj}
\end{equation}
where $\beta$ controls the information bottleneck of the compressed latent space.

\subsection{Conditional Latent Flow Matching}
\label{subsec:flow_matching}

While diffusion models are often computationally demanding \cite{diffslow}, flow matching \cite{flowmatching} offers a faster alternative by learning a time-dependent vector field $v_\theta$ that pushes a simple noise distribution $p_0$ toward the data distribution $p_1$. We define a linear probability path $\psi_t(z)$ between noise $z_0 \sim \mathcal{N}(0, I)$ and a compressed latent $z_1$ as:
\begin{equation}
    z_t = \psi_t(z_0) = (1-t)z_0 + t z_1, \quad t \in [0, 1].
\end{equation}

The neural network $v_\theta(z_t, t, f)$ is trained to approximate the conditional velocity field $v_t(z_t\mid z_1) = z_1 - z_0$ by minimizing the \gls{cfm} objective:
\begin{equation}
\label{eq:cfm}
    \mathcal{L}_{\text{CFM}} = \mathbb{E}_{t, z_0, z_1} \Big[ \big\| v_{\theta}(z_t, t, f) - (z_1 - z_0) \big\|^2 \Big],
\end{equation}
where $f$ denotes the additional fitness conditioning. At inference, new candidates are generated by numerically integrating the learned velocity field:
\begin{equation}
    z_{1} = z_0 + \int_0^1 v_{\theta}(z_t, t, f) \, dt.
\end{equation}
In practice, this integration is performed using a first-order Euler solver with $K$ discrete steps of size $\Delta t = 1/K$. 
The resulting latent $z_{1}$ is then mapped back to sequence space via the frozen decompressor and decoder, \mbox{$x' = \mathcal{D}(\mathcal{R}(z_{1}))$}.

To enable fitness-conditioned generation without using an external predictor model, we parameterize $v_\theta$ as a U-Net and train it using \gls{cfg}~\cite{classfreeguidence}. During training, the fitness condition $f$ is randomly dropped with probability $p$, allowing the model to simultaneously learn the conditional distribution $v_{\text{cond}} = v_\theta(z_t, t, f)$ and the unconditional distribution $v_{\text{uncond}} = v_\theta(z_t, t, \emptyset)$. During sampling, the final velocity field $\epsilon$ is computed as a weighted combination of these two predictions:
\begin{equation}
    \epsilon = (1+w) \cdot v_{\text{cond}} - w \cdot v_{\text{uncond}},
\end{equation}
where $w$ is the guidance scale. 
This formulation enables efficient steering toward high-fitness regions, where an external predictor is used only to set a suitable target fitness $f$, obviating the need for it during the actual sampling steps.
Apart from one additional forward pass when $w \neq 0$, this method is no more costly than basic sampling, making it significantly faster than methods requiring to evaluate an additional likelihood term. 
The parameters $\theta$ of the flow matching model are optimized while all \gls{vae} components remain frozen. 

\subsection{Bootstrapping and Data Augmentation}
To address the inherent sparsity and imbalance of sequence-fitness landscapes towards low-fitness regimes (see \cref{tab:datasets}), we use a bootstrapping strategy to enrich the training signal. Our goal is to augment the initial limited dataset $\mathcal{A}$ with synthetic variants that uniformly populate a target fitness interval $\mathcal{I}$, thereby improving the coverage of the generative model.

Using the flow matching model $v_{\theta}$ trained on the original data $\mathcal{A}$, we select a set of target fitness values $\mathcal{F}$ spaced evenly across $\mathcal{I}$. We expand the dataset by a fixed factor (we set this to $25\%$), allocating the generation budget evenly across $\mathcal{F}$. For each target value $f \in \mathcal{F}$, we generate a batch of synthetic sequences $x$ using the guided sampling procedure in \cref{alg:data_augmentation}. The associated labels $\hat{f}$ are perturbed with Gaussian noise to account for the potential discrepancy between the target fitness and the actual fitness of the generated sequence—a mismatch expected given that $v_\theta$ is initially trained on limited data:
\begin{equation}
    \hat{f} = f + q \cdot \eta, \qquad \eta \sim \mathcal{N}(0, 1),
\end{equation}
where $q$ is a dataset-specific scale factor. This perturbation ensures that the augmented labels represent a realistic distribution around the target values rather than an over-confident point estimate. 
The final augmented dataset $\mathcal{A}_{\text{aug}} = \mathcal{A} \cup \mathcal{A}_{\text{syn}}$ is then used to retrain the flow model $v_{\theta}$ (the other components remain frozen), again employing score dropout rate during training as regularization due to the additional synthetic labels. The specific interval bounds and hyperparameters for each benchmark are detailed in the Appendix, \cref{sec:appendix_exp_details}.

\begin{algorithm}[ht]
\caption{Data Augmentation via Guided Sampling}
\label{alg:data_augmentation}
\begin{algorithmic}[1]
\REQUIRE $K,\ w,\ \mathcal{F},\ q,\ v_\theta$
\STATE Set step size $\Delta t \gets \frac{1}{K}$
\STATE $\mathcal{A}_{\text{syn}} \gets \emptyset$
\FOR{$f \in \mathcal{F}$}
    \STATE Initialize $z_0 \sim \mathcal{N}(0, \text{I})$ 
    \FOR{$k = 0$ to $K-1$}
        \STATE $t \gets k \cdot \Delta t$
        \STATE $v_{\text{cond}} \gets v_\theta(z_t, t, f)$
        \STATE $v_{\text{uncond}} \gets v_\theta(z_t, t, \emptyset)$
        \STATE $\epsilon \gets (1+w)\cdot v_{\text{cond}} - w \cdot v_{\text{uncond}}$
        \STATE $z_{t+\Delta t} \gets z_t + \Delta t \cdot \epsilon$
    \ENDFOR
    \STATE $\eta \sim \mathcal{N}(0, 1)$
    \STATE $x \gets \mathcal{D}(\mathcal{R}(z_K))$
    \STATE $\mathcal{A}_{\text{syn}} \gets \mathcal{A}_{\text{syn}} \cup \{ (x, f + q \cdot \eta) \}$ 
\ENDFOR
\STATE $\mathcal{A}_{\text{aug}} \gets \mathcal{A} \cup \mathcal{A}_{\text{syn}}$
\end{algorithmic}
\end{algorithm}


\section{Experiments}
\label{sec:experiments}

\subsection{Datasets}
We evaluate our approach using the four protein fitness optimization benchmarks introduced in \cite{smoothgwg}: the {Medium} and {Hard} variants of the \gls{aav}~\cite{aav} and \gls{gfp}~\cite{gfp} datasets. These tasks represent low-data scenarios where the model must extrapolate beyond the training distribution.

Each benchmark is a restricted subset of a full ground-truth dataset $\mathcal{A}_{\text{GT}}$, where difficulty is determined by the quality of initial sequences and their mutational distance to the sequences with the highest fitness in the ground-truth data set $\mathcal{A}_{\text{GT}}$. In the Medium tasks, training sequences are sampled from the $20^{\text{th}}$--$40^{\text{th}}$ percentile of $\mathcal{A}_{\text{GT}}$ and require at least 6 mutations to reach the top $1\%$ of the landscape. The Hard tasks are more constrained, restricting sequences to those below the $30^{\text{th}}$ percentile and requiring a minimum distance of 7 mutations to achieve the same high-fitness region.

As shown in \cref{tab:datasets}, these restrictions result in highly imbalanced datasets with narrow fitness ranges (e.g., AAV Medium only covers $[0.29, 0.38]$). This data scarcity motivates our use of latent manifold construction and bootstrapping to effectively navigate the landscape. All fitness scores are normalized to $[0, 1]$ based on the extrema of the full dataset $\mathcal{A}_{\text{GT}}$.

For the \textit{in-silico} evaluation of generated sequences, we utilize two pretrained fitness oracles $\Omega_{\phi}$ (one for \gls{aav} and one for \gls{gfp}). These oracles are taken directly from \cite{smoothgwg}, as they serve as the standard evaluation proxy for all methods within this benchmark. Each oracle was trained on the complete ground-truth dataset $\mathcal{A}_{\text{GT}}$, ensuring a reliable and consistent measure of biological performance during the testing phase.

\begin{table}[h]
\centering
\caption{Characteristics of ground-truth and protein optimization benchmark datasets. Fitness ranges are normalized.}
\begin{tabular}{lccc}
    \toprule
        \textbf{Dataset} & \textbf{Num. Sequences} & \textbf{Fitness Range} \\
        \midrule
        AAV & 44,128 & $[0.0, 1.0]$ \\
        GFP & 56,086 & $[0.0, 1.0]$ \\
        \midrule
        AAV Medium & 2,139 & $[0.29, 0.38]$ \\
        AAV Hard & 3,448 & $[0.0, 0.33]$ \\
        GFP Medium & 2,828 & $[0.01, 0.62]$ \\
        GFP Hard & 2,426 & $[0.0, 0.10]$ \\
        \bottomrule
\end{tabular}
\label{tab:datasets}
\end{table}

\subsection{Metrics}
Following standard protocol of the used benchmark \cite{smoothgwg}, all generated sequences are evaluated post-hoc using the pretrained fitness oracle $\Omega_{\phi}$. For each model, we perform five independent runs using different random seeds. In each run, we sample 512 latent points to generate a set of sequences $S_i$.  

To maintain consistency with the established benchmark protocol \cite{smoothgwg}, we rank the remaining candidates using the same ranking predictor as the baseline methods taken from \cite{smoothgwg}, selecting the top-$k$ ($k=128$) sequences for final evaluation in $S_i$. It is important to note that while these predictors are used for post-hoc ranking to facilitate comparison, the generative sampling process itself remains independent of them.

We evaluate the generated sequences across three metrics, as defined in \cite{smoothgwg, gflow} (see \cref{sec:metrics} in the Appendix for detailed formulation): fitness $f$, which represents the median score predicted by the oracle $\Omega_{\phi}$; diversity, calculated as the average pairwise distance between sequences in the generated set $S_i$; and novelty, computed as the average distance between sequences in $S_i$ and their nearest training set neighbors.

We report the mean $\mu$ and standard deviation $\sigma$ over the five independent runs. It is important to note that $\Omega_{\phi}$ is used exclusively for evaluation; it is never involved in training the flow model, guiding the sampling process, or generating synthetic data. This strict separation ensures an unbiased assessment of the model's ability to extrapolate from the restricted training set $\mathcal{A}$ towards high-fitness regions of the global landscape.

\subsection{Results}

\begin{table*}[ht]
\vskip 0.15in
\caption{AAV model performance comparison, best fitness values are \textbf{bold} and second best are \underline{underlined}.}
\vskip 0.15in
\begin{center}
\begin{tabular}{lccc|ccc}
\toprule
 &\multicolumn{3}{c|}{\textbf{AAV Medium}} & \multicolumn{3}{c}{\textbf{AAV Hard}}  \\
         \textbf{Method} & \textbf{Fitness}$\uparrow$ & {Diversity} & {Novelty} & \textbf{Fitness} $\uparrow$ & {Diversity} & {Novelty}\\
       \toprule
        GFN-AL & 0.20{ \scriptsize ± 0.1} & ~~~~9.6{ \scriptsize ± 1.2} & 19.4{ \scriptsize ± 1.1} & 0.10{ \scriptsize ± 0.1} & 11.6{ \scriptsize ± 1.4} & 19.6{ \scriptsize ± 1.1} \\ 
        CbAS & 0.43{ \scriptsize ± 0.0} & ~~12.7{ \scriptsize ± 0.7} & ~~7.2{ \scriptsize ± 0.4} & 0.36{ \scriptsize ± 0.0} & 14.4{ \scriptsize ± 0.7} & ~~8.6{ \scriptsize ± 0.5} \\
        AdaLead & 0.46{ \scriptsize ± 0.0} & ~~~~8.5{ \scriptsize ± 0.8} & ~~2.8{ \scriptsize ± 0.4} & 0.40{ \scriptsize ± 0.0} & ~~8.5{ \scriptsize ± 0.1} & ~~3.4{ \scriptsize ± 0.5} \\
        BOqei & 0.38{ \scriptsize ± 0.0} & ~~15.2{ \scriptsize ± 0.8}  & ~~0.0{ \scriptsize ± 0.0} & 0.32{ \scriptsize ± 0.0} & 17.9{ \scriptsize ± 0.3} & ~~0.0{ \scriptsize ± 0.0} \\
        CoMS & 0.37{ \scriptsize ± 0.1} & ~~10.1{ \scriptsize ± 5.9} & ~~8.2{ \scriptsize ± 3.5} & 0.26{ \scriptsize ± 0.0} & 10.7{ \scriptsize ± 3.5} & 10.0{ \scriptsize ± 2.8} \\
        PEX & 0.40{ \scriptsize ± 0.0} & ~~~~2.8{ \scriptsize ± 0.0} & ~~1.4{ \scriptsize ± 0.2} & 0.30{ \scriptsize ± 0.0} & ~~2.8{ \scriptsize ± 0.0} & ~~1.3{ \scriptsize ± 0.3} \\
        \gls{gpe}* & 0.53{ \scriptsize ± 0.0} & ~~~~5.2{ \scriptsize ± 0.2} & ~~5.6{ \scriptsize ± 0.6} & 0.54{ \scriptsize ± 0.0} & ~~4.6{ \scriptsize ± 0.7} & ~~6.6{ \scriptsize ± 0.5} \\
        \midrule
        gg-dWJS & 0.48{ \scriptsize ± 0.0} & ~~~~9.5{ \scriptsize ± 0.3} & ~~4.2{ \scriptsize ± 0.4} & 0.33{ \scriptsize ± 0.0} & 14.3{ \scriptsize ± 0.7} & ~~5.3{ \scriptsize ± 0.4} \\
        LatProtRL & 0.57{ \scriptsize ± 0.0} & ~~~~$\!$3.0{ \scriptsize {± n.a.}} & ~~$\!$5.0{ \scriptsize {± n.a.}} & 0.57{ \scriptsize ± 0.0} & ~~~$\!$3.0{ \scriptsize {± n.a.}} & ~~~$\!$7.0{ \scriptsize {± n.a.}} \\
        GWG & 0.43{ \scriptsize ± 0.1} & ~~~6.6{ \scriptsize ± 6.3} &  ~7.7{ \scriptsize ± 0.8} & 0.33{ \scriptsize ± 0.0} & 12.0{ \scriptsize ± 0.4} & 12.2{ \scriptsize ± 0.4} \\
        GGS & 0.51{ \scriptsize ± 0.0} & ~~~4.0{ \scriptsize ± 0.2} &~5.4{ \scriptsize ± 0.5} & \underline{0.60}{ \scriptsize ± 0.0} & ~~4.5{ \scriptsize ± 0.5} & ~~7.0{ \scriptsize ± 0.0} \\
        VLGPO unsmoothed  & \underline{0.58}{ \scriptsize ± 0.0} & ~~~5.6{ \scriptsize ± 0.2} & ~5.0{ \scriptsize ± 0.0} & 0.51{ \scriptsize ± 0.0} & ~~8.4{ \scriptsize ± 0.2} & ~~7.8{ \scriptsize ± 0.4}  \\
        VLGPO smoothed  & 0.53{ \scriptsize ± 0.0} & ~~~5.0{ \scriptsize ± 0.2} & ~5.0{ \scriptsize ± 0.0} &\textbf{0.61}{ \scriptsize ± 0.0} & ~~4.3{ \scriptsize ± 0.1} & ~~6.2{ \scriptsize ± 0.4} \\
        \midrule
        \gls{method} & \textbf{0.62}{ \scriptsize ± 0.0} & ~~~4.7{ \scriptsize ± 0.1} & ~6.0{ \scriptsize ± 0.0} & \textbf{0.61}{ \scriptsize ± 0.0} & ~~4.4{ \scriptsize ± 0.1}& ~~7.0{ \scriptsize ± 0.0}\\
        \bottomrule
\end{tabular}
\label{tab:aav_res}
\end{center}
\end{table*}
\begin{table*}
\begin{center}
\caption{GFP model performance comparison, best fitness values are \textbf{bold} and second best are \underline{underlined}.}
\begin{tabular}{lccc|ccc}
\toprule
 &\multicolumn{3}{c|}{\textbf{GFP Medium}} & \multicolumn{3}{c}{\textbf{GFP Hard}}  \\
         \textbf{Method} & \textbf{Fitness}$\uparrow$ & {Diversity} & {Novelty} & \textbf{Fitness} $\uparrow$ & {Diversity} & {Novelty}\\
       \toprule
       GFN-AL & 0.09{ \scriptsize  ± 0.1} & 25.1{ \scriptsize  ± 0.5} & ~213{ \scriptsize  ± 2.2} & ~~0.1{ \scriptsize  ± 0.2} & ~23.6{ \scriptsize  ± 1.0} & ~214{ \scriptsize  ± 4.2} \\ 
        CbAS & 0.14{ \scriptsize  ± 0.0} & ~~9.7{ \scriptsize  ± 1.1} & ~~7.2{ \scriptsize  ± 0.4} & 0.18{ \scriptsize  ± 0.0} & ~~~9.6{ \scriptsize  ± 1.3} & ~~7.8{ \scriptsize  ± 0.4} \\
        AdaLead & 0.56{ \scriptsize  ± 0.0} & ~~3.5{ \scriptsize  ± 0.1} & ~~2.0{ \scriptsize  ± 0.0} & 0.18{ \scriptsize  ± 0.0} & ~~~5.6{ \scriptsize  ± 0.5} & ~~2.8{ \scriptsize  ± 0.4} \\
        BOqei & 0.20{ \scriptsize  ± 0.0} & 19.3{ \scriptsize  ± 0.0} & ~~0.0{ \scriptsize  ± 0.0} & ~~0.0{ \scriptsize  ± 0.5} & $\,$94.6{ \scriptsize  ± 71} & ~~54{ \scriptsize  ± 81} \\
        CoMS & 0.00{ \scriptsize  ± 0.1} & 133{ \scriptsize  ± 25} & 192{ \scriptsize  ± 12} & ~~0.0{ \scriptsize  ± 0.1} & ~~144{ \scriptsize  ± 7.5} & ~201{ \scriptsize  ± 3.0} \\
        PEX & 0.47{ \scriptsize  ± 0.0} & ~~3.0{ \scriptsize  ± 0.0} & ~~1.4{ \scriptsize  ± 0.2} & ~~0.0{ \scriptsize  ± 0.0} & ~~~3.0{ \scriptsize  ± 0.0} & ~~1.3{ \scriptsize  ± 0.3} \\
        \gls{gpe}* &  0.84{ \scriptsize  ± 0.0} & ~~1.9{ \scriptsize  ± 0.2} & ~~7.0{ \scriptsize  ± 0.7} & \underline{0.88}{ \scriptsize  ± 0.0} & ~~~2.8{ \scriptsize  ± 0.2} & ~~7.0{ \scriptsize  ± 0.0} \\
        \midrule
        gg-dWJS & 0.55{ \scriptsize  ± 0.1} & 52.3{ \scriptsize  ± 3.4} & 16.3{ \scriptsize  ± 5.7} & 0.61{ \scriptsize  ± 0.1} & ~68.0{ \scriptsize  ± 5.6} & 44.8{ \scriptsize  ± 47~$\,$} \\
        LatProtRL & 0.81{ \scriptsize  ± 0.0} &~~~$\!$3.0{ \scriptsize {± n.a.}} & ~~~$\!$5.0{ \scriptsize {± n.a.}} & 0.75{ \scriptsize  ± 0.0} & ~~~~$\!$3.0{ \scriptsize {± n.a.}} & ~~~$\!$7.0{ \scriptsize {± n.a.}} \\
        GWG & 0.10{ \scriptsize  ± 0.0} &33.0{ \scriptsize  ± 0.8} &12.8{ \scriptsize  ± 0.4} & ~~0.0{ \scriptsize  ± 0.0} & ~~~4.2{ \scriptsize  ± 7.0} &~~7.6{ \scriptsize  ± 1.1} \\
        GGS & 0.76{ \scriptsize  ± 0.0} &~~3.7{ \scriptsize  ± 0.2} &~~5.0{ \scriptsize  ± 0.0} & 0.74{ \scriptsize  ± 0.0} & ~~~3.6{ \scriptsize  ± 0.1} & ~~8.0{ \scriptsize  ± 0.0} \\
        VLGPO unsmoothed  & \underline{0.87}{ \scriptsize  ± 0.0} & ~~4.3{ \scriptsize  ± 0.1} & ~~6.0{ \scriptsize  ± 0.0} & 0.75{ \scriptsize  ± 0.0} & ~~~3.1{ \scriptsize  ± 0.2} & ~~6.0{ \scriptsize  ± 0.0}  \\
        VLGPO smoothed  & 0.84{ \scriptsize  ± 0.0} & ~~2.1{ \scriptsize  ± 0.1} & ~~5.0{ \scriptsize  ± 0.0} & 0.78{ \scriptsize  ± 0.0} & ~~~2.5{ \scriptsize  ± 0.2} & ~~6.0{ \scriptsize  ± 0.0} \\
        \midrule
        \gls{method} & \textbf{0.91}{ \scriptsize  ± 0.0}& ~~~4.1 {\scriptsize  ± 0.1} &~~8.0{ \scriptsize  ± 0.0} &\bfseries{0.92}{ \scriptsize  ± 0.0}& ~~~1.1{ \scriptsize  ± 0.2} & ~~6.8{ \scriptsize  ± 0.2} \\
        \bottomrule
    \end{tabular}
\label{tab:gfp_res}
\end{center}
\end{table*}

We evaluate \gls{method} on the four benchmarks of \gls{aav} Medium, \gls{aav} Hard, \gls{gfp} Medium, and \gls{gfp} Hard, comparing its performance against several baseline methods. This includes several model-based and adaptive sampling approaches, such as GFlowNet (GFN-AL) \cite{gflow}, model-based adaptive sampling (CbAS) \cite{brookes}, greedy search (AdaLead) \cite{adalead}, Bayesian optimization (BOqei)~\cite{wilson2017reparameterization}, conservative model-based optimization (CoMS)~\cite{trabucco2021conservative}, proximal exploration (PEX) \cite{ren}, and \gls{gpe} \cite{pinheiroimplicit}. Since \gls{gpe} reports results for four different setups we choose their best candidates for each benchmark (denoted as \gls{gpe}*), since there was not one setting which was consistently outperforming.  
Furthermore, we compare against a class of methods that rely on predictor-guidance during the sampling process, including gg-dWJS \cite{gradguidancemanifold}, \gls{gwg} \cite{oops}, its smoothed variant \gls{ggs} \cite{smoothgwg}, and LatProtRL \cite{lee2024robust}. Finally, we include results for \gls{vlgpo} \cite{vlgpo} using both smoothed and unsmoothed predictors.

Following prior work, we treat fitness as the primary metric and report diversity and novelty as secondary metrics. This is due to the fact that the latter metrics are not intended to be maximized indefinitely; excessively high values would indicate a pure exploration search and could be achieved by entirely random sequences that fail to satisfy the fitness enhancement criterion. The complete quantitative results for \gls{aav} and \gls{gfp} are provided in \cref{tab:aav_res,tab:gfp_res}. Methods listed in the second section of \cref{tab:aav_res,tab:gfp_res} utilize the same external fitness predictor (trained on $\mathcal{A}$), either as a source of gradient-guidance or as a reward function~\cite{smoothgwg}.

We obtain state-of-the-art fitness metrics on all four benchmarks. While diversity values generally remain within ranges comparable to other baseline methods, the novelty (defined as the minimum distance between a generated sequence and any sequence in the training dataset $\mathcal{A}$ as a measure of extrapolation) is even slightly increased relative to most competitive baseline methods.

While gradient-based sampling yields strong fitness scores, especially in \gls{vlgpo} and \gls{ggs}, it also introduces undesirable side effects. In particular, high-fitness samples tend to collapse to a small number of unique sequences, reducing diversity, which is seen more clearly with the smoothed predictor. Our conditional sampling approach removes the need of an external predictor for guiding the sampling process, thereby allowing for a more direct exploration of the fitness landscape. This shift is reflected in our experimental results, where we note that an increase is fitness at the cost of decreased diversity is not always the case for \gls{method}.
This can be observed for \gls{aav} Hard, where state-of-the-art fitness is joined by improved diversity values compared to \gls{vlgpo} smoothed and \gls{ggs}, which both show similar fitness but decreased diversity. Similarly, for \gls{aav} Medium, \gls{method} achieves gains in fitness while maintaining competitive diversity levels.
The relationship between fitness, diversity, and novelty is further illustrated in \cref{fig:pareto} for \gls{gfp} Medium. These results demonstrate that \gls{method} achieves superior fitness compared to existing baselines while maintaining competitive diversity and novelty. 

We attribute these performance gains to the enhanced latent representation of \gls{method}, which is based on the rich protein sequence embeddings captured by \glspl{plm}. Methods such as \gls{gwg}, \gls{ggs}, and gg-dWJS operate directly on the discrete sequence space or the one-hot encoded sequence space, respectively, which may become increasingly challenging as the sequence space grows. In contrast, while \gls{vlgpo} utilizes a learned continuous latent space, its reliance on a trained \gls{vae} using a small dataset of protein family does not exploit the existing knowledge within pretrained protein sequence representations. Furthermore, although LatProtRL also operates within a \gls{plm} embedding space, it is framed as an online optimization problem, meaning we can only compare the results of the initial rounds. Moreover, it does not employ a generative model to capture and exploit the latent representation—working instead with uncompressed embeddings—a limitation that is likely reflected in the decreased diversity observed across all four benchmarks.

Beyond performance gains, a key practical advantage of \gls{method} is that competitive results are obtained without the need for gradient-based inference during the \gls{ode} sampling steps. By directly sampling from the learned fitness-conditioned distribution, we eliminate time-expensive predictor guidance steps, allowing for efficient and fast generation of high-fitness variants. 
We therefore benchmark \gls{method} against both variants of \gls{vlgpo} as the most competitive baseline. The results in \cref{table:runtime_memory} show that \gls{method} reduces inference costs by factors of 10 to 85 compared to \gls{vlgpo}. 
While \gls{method} exhibits a 2- to 3-fold increase in memory consumption, its absolute memory usage remains low, thus representing a favorable trade-off for a substantial gain in efficiency with much lower inference times.

\subsection{Bootstrapping and Data Augmentation}

\begin{table}[!ht]
    \caption{Bootstrapping results for all four tasks on \gls{aav}/\gls{gfp} with medium and hard difficulty.}
    \label{table:bootstrapping}
    \centering
    \resizebox{.99\linewidth}{!}{
    \begin{tabular}{lcccc}
        \toprule
        & \gls{aav} Medium & \gls{aav} Hard &  \gls{gfp} Medium & \gls{gfp} Hard \\
        Method & Fitness~$\mathrm{\uparrow}$ & Fitness~$\mathrm{\uparrow}$ & Fitness~$\mathrm{\uparrow}$ & Fitness~$\mathrm{\uparrow}$ \\
        \midrule 
        \gls{method} & 0.62{ \footnotesize ± 0.00} & 0.61{ \footnotesize ± 0.00}  & 0.91{ \footnotesize ± 0.01} & \textbf{0.92}{ \footnotesize ± 0.01}\\
        \gls{method} Bootstrapped & \textbf{0.65}{ \footnotesize ± 0.00}  & ~\textbf{0.63}{ \footnotesize ± 0.00 } & \textbf{0.93}{ \footnotesize ± 0.01 }& 0.87{ \footnotesize ± 0.01} \\
        \bottomrule
    \end{tabular}
    }
    \vspace{0.5em}
\end{table}

The results for applying our proposed bootstrapping scheme on the four benchmarks are shown in \cref{table:bootstrapping}. We observe additional gains in fitness values on benchmarks where there is still significant room for improvement (in particular \gls{aav} Medium and Hard).
This indicates that augmenting the limited initial dataset with samples generated by the model itself can further boost performance, likely by exploiting both the interpolation and extrapolation capabilities of our approach. In contrast, no additional gains are observed for the \gls{gfp} Hard task; we hypothesize that the synthetic data for this specific landscape may be too noisy to provide a clear signal for further optimization.
While there is inherent uncertainty in the fitness labels assigned to the bootstrapped dataset $\mathcal{A}_{\text{syn}}$, the results suggest that \gls{method} effectively extracts meaningful signals from these synthetic samples. Specifically, the model appears to learn the approximate ranking of sequences within the fitness landscape without becoming overly sensitive to the absolute values of the synthetic labels, which may be inexact.

\subsection{Ablations}

We performed several ablation studies to assess how architectural and training choices influence model performance. First, we examined the impact of the compressor’s factor $c$ by comparing our primary setting ($c=20$) with a less aggressive compression of $c=16$. As shown in \cref{table:ablation_cf}, $c=20$ yielded superior results, suggesting that compressing the \gls{plm} latent space acts as a beneficial bottleneck that encourages the model to retain the most important features. 

Moreover, we compared two \gls{vae} training strategies to evaluate their impact on generative quality. Our primary setup employs a two-stage training approach: the decoder is first pre-trained to reconstruct \gls{plm} embeddings, followed by a full encoder-decoder fine-tuning phase. We contrast this with a one-stage training schedule involving joint optimization of the decoder, compressor and decompressor from scratch.
Performance metrics for both strategies are detailed in \cref{table:ablation_2stage}. While results for the \gls{aav} Medium dataset are comparable across both regimes, the two-stage training strategy generally yields superior performance in the final generated samples. This suggests that stabilizing the decoder's representation of the \gls{plm} space prior to learning the compression is more robust. 
Additional ablations on the employed score dropout $p$ during flow matching training can be found in~\cref{sec:additional_ablations} (Appendix).

\begin{table}[!ht]
    \caption{Influence of compression factor for all four tasks on \gls{aav}/\gls{gfp} with medium and hard difficulty.}
    \label{table:ablation_cf}
    \centering
    \resizebox{.99\linewidth}{!}{
    \begin{tabular}{lcccc}
        \toprule
        & \gls{aav} Medium & \gls{aav} Hard &  \gls{gfp} Medium & \gls{gfp} Hard \\
        Method & Fitness~$\mathrm{\uparrow}$ & Fitness~$\mathrm{\uparrow}$ & Fitness~$\mathrm{\uparrow}$ & Fitness~$\mathrm{\uparrow}$ \\
        \midrule 
        \gls{method}/20 & \textbf{0.62}{ \footnotesize ± 0.00} & \textbf{0.61}{ \footnotesize ± 0.00}  & \textbf{0.91}{ \footnotesize ± 0.01} & \textbf{0.92}{ \footnotesize ± 0.01}\\
        \gls{method}/16 & 0.53{ \footnotesize ±  0.01} & 0.53{ \footnotesize ± 0.01}  & 0.87{ \footnotesize ± 0.02} & 0.88{ \footnotesize ± 0.02} \\
        \bottomrule
    \end{tabular}
    }
    \vspace{0.5em}
\end{table}

\begin{table}[!ht]
    \caption{Influence of two-stage training on resulting fitness for all four tasks on \gls{aav}/\gls{gfp} with medium and hard difficulty.}
    \label{table:ablation_2stage}
    \centering
    \resizebox{.99\linewidth}{!}{
    \begin{tabular}{lcccc}
        \toprule
        & \gls{aav} Medium & \gls{aav} Hard &  \gls{gfp} Medium & \gls{gfp} Hard \\
        Method & Fitness~$\mathrm{\uparrow}$ & Fitness~$\mathrm{\uparrow}$ & Fitness~$\mathrm{\uparrow}$ & Fitness~$\mathrm{\uparrow}$ \\
        \midrule 
        \gls{method} w/ 2-Stage & \textbf{0.62}{ \footnotesize ± 0.00} & \textbf{0.61}{ \footnotesize ± 0.00}  & \textbf{0.91}{ \footnotesize ± 0.01} & \textbf{0.92}{ \footnotesize ± 0.01}\\
        \gls{method} w/ 1-Stage & \textbf{0.62}{ \footnotesize ± 0.01} & 0.53{ \footnotesize ± 0.01} &0.89{ \footnotesize ± 0.01} & 0.88{ \footnotesize ± 0.04} \\
        \bottomrule
    \end{tabular}
    }
    \vspace{0.5em}
\end{table}

\section{Discussion}
\label{sec:discussion}

In this work, we introduced a method for protein fitness optimization that leverages the compressed embedding space of a pretrained \gls{plm} and fitness-conditioned flow matching to model the conditional latent distribution. By utilizing the expressive latent representations of the \gls{plm}, our approach enables rapid conditional sampling and the efficient generation of high-fitness protein variants without requiring backpropagation through fitness predictor models for guidance. 
Despite its simplicity, the proposed method achieves state-of-the-art performance on established protein optimization benchmarks. Furthermore, we demonstrated that our framework can be bootstrapped: by generating and incorporating synthetic data, a retrained flow model can outperform its predecessor in most of the benchmarks, offering a promising strategy to enrich low-data protein regimes.

While our sampling process is predictor-free—in that it does not require external gradient guidance during generation—evaluating against standard benchmarks still necessitates ranking the top-$k$ candidates via an external model. Beyond evaluation, the predictor played a role in hyperparameter selection, specifically in calibrating the target fitness f to balance the model’s ability to extrapolate toward higher fitness values. Future research could explore distilling the fitness predictor directly into the flow model to move toward a truly end-to-end generative pipeline.

Moreover, the focus of this evaluation on \gls{aav} and \gls{gfp} benchmarks—consistent with current community standards~\cite{smoothgwg,lee2024robust,pinheiroimplicit,vlgpo}—represents a scope limitation that should be addressed in future research. From a conceptual standpoint, \gls{method}—as well as the baseline approaches considered—is agnostic to the specific protein target and can be readily extended to a broader range of landscapes, such as those curated in FLIP~\cite{dallago2021flip} or ProteinGym~\cite{notin2023proteingym}. Extending the framework to these more diverse datasets will be essential to fully characterize the model's robustness across functional domains.

Additionally, investigating alternative sampling strategies, such as temperature-based control \cite{temp} or the integration of a repulsion term \cite{balcerak2025energy}, may further enhance the diversity of the generated sequences. Finally, integrating active online feedback from a fitness predictor or employing uncertainty-aware objectives could further boost performance and improve fitness accuracy, particularly in scenarios where experimental data is limited.

\section*{Acknowledgements}
This work was supported by the University Research Priority Program (URPP) Human Reproduction Reloaded of the University of Zurich.

\section*{Impact Statement}
This paper presents a computational method for protein fitness optimization aimed at advancing the search for proteins that cure diseases and reduce side effects. However, we acknowledge that fitness scores can be defined arbitrarily and could, in theory, be misused for unethical purposes.

\bibliography{references}
\bibliographystyle{icml2026}

\newpage

\appendix
\onecolumn

\section{Detailed Architecture}

\begin{figure*}[h]
    \centering
    \includegraphics[width=0.9\linewidth]{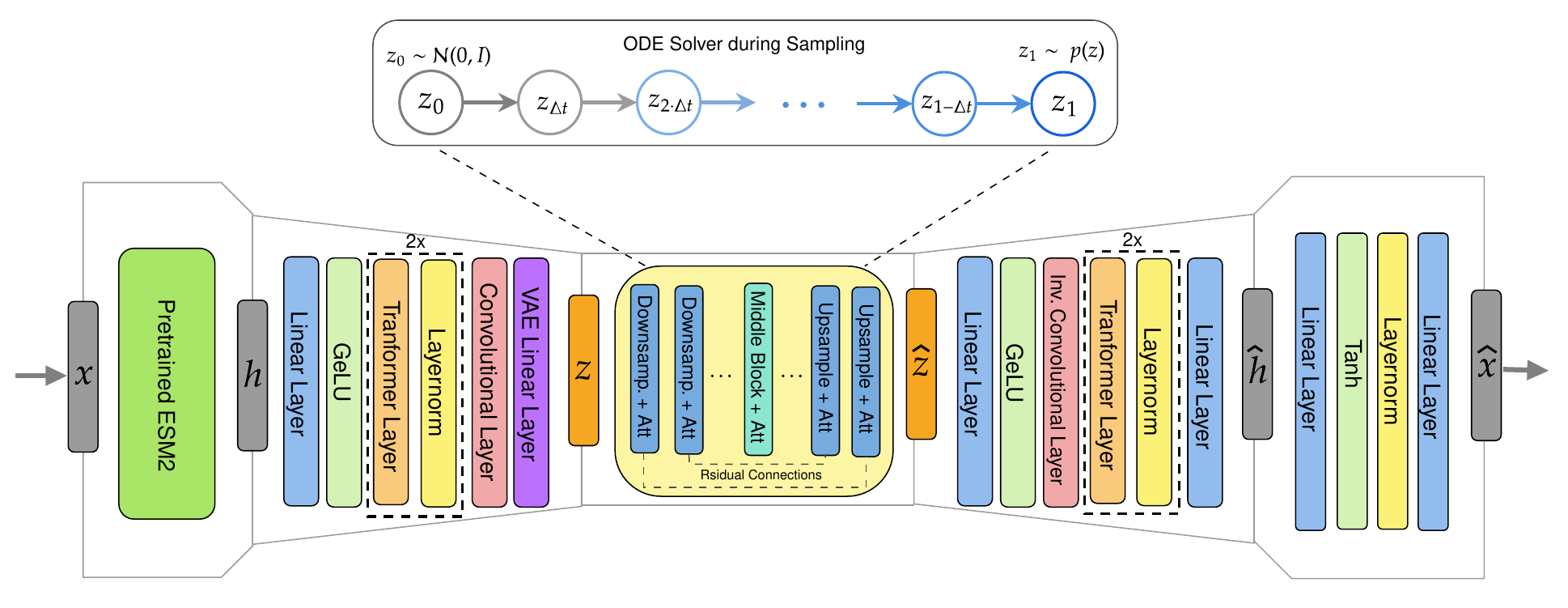}
    \caption{Detailed version of our architecture for the encoder/decoder setup, the compressor/decompressor and the flow matching model. The addition of the conditional score $f$ and the time conditioning $t$ was omitted for brevity.}
    \label{fig:detailed_architecture}
\end{figure*}

A detailed schematic of the employed architecture is presented in \cref{fig:detailed_architecture}. The encoder/decoder and compressor/decompressor configurations are based on the architecture described in \cite{protflow}. For the conditional flow matching component, we utilize a U-Net implementation adapted from a widely-used open-source repository.\footnote{\url{https://github.com/lucidrains/denoising-diffusion-pytorch}}

\section{Experimental Details and Hyperparameters}
\label{sec:appendix_exp_details}

We use an 80/20 train-test split for all benchmark datasets. A random split was employed because the multi-positional nature of the mutants makes it difficult to define independent folds, as discussed in~\cite{notin2023proteingym}. All models are optimized using the AdamW optimizer with default hyperparameters, with the exception of the specific learning rates detailed in the following sections.

\subsection{VAE Training}
We primarily adopt a two-stage training procedure for the autoencoding backbone. In the first stage, the decoder is trained while keeping the pretrained \gls{plm} encoder frozen. We use a batch size of 128, a learning rate of $5\cdot 10^{-5}$, and 200 warm-up steps. Training continues using cross-entropy loss until convergence (maximum 30 epochs) with a patience of 8, evaluated every 50 steps.

In the second stage, we train the compressor/decompressor pair while keeping both the encoder and decoder frozen. This stage utilizes the combined objective in \cref{eq:comb_obj} with $\beta=10^{-4}$. We train for a maximum of 400 epochs with a learning rate of $5\cdot 10^{-4}$, a batch size of 128, and a patience of 8, evaluating every 400 steps. 

As an ablation study, we also evaluated a joint training approach where these three components, the decoder as well as the compressor/decompressor are optimized simultaneously using the parameters from the second stage, see~\cref{table:ablation_2stage}. For latent compression, we evaluated factors of $c \in \{16, 20\}$, denoted as \gls{method}/16 and \gls{method}/20 in our results, see~\cref{table:ablation_cf}.

\subsection{Flow Matching and Sampling}
The flow matching model $v_{\theta}$ is trained using the objective $\mathcal{L}_{\text{CFM}} = \mathcal{L}_{\text{MSE}}(v_t(z_t), v_{\theta}(z_t, t, f))$. We use a learning rate of $2\cdot 10^{-4}$, a warm-up of 400 steps, and a batch size of 256. The U-Net architecture utilizes 2 down-scaling and up-scaling blocks. We train for 800k steps for \gls{aav} and 600k and 500k steps for \gls{gfp} Medium and Hard, respectively. 
The score dropout rate $p$ is set to $\{0.0, 0.2, 0.0, 0.0\}$ for \gls{aav} Medium and Hard, and \gls{gfp} Medium and Hard, respectively. 
For conditional sampling, we fix the number of ODE steps to $K=40$, but we did not find any notable differences when this number was varied.
Target fitness values were set to $\{0.5, 0.55, 0.8, 1.4\}$, which were obtained using the pretrained predictor model from~\cite{smoothgwg}, as well as the \gls{cfg} parameter $w$, which was set to $\{0.2, 0.1, -0.08, 0.1\}$.

\subsection{Bootstrapping Hyperparameters}
The target fitness intervals $\mathcal{I}$ were chosen to maintain a margin above the minimum score while remaining below known top-performing variants, as indicated by the available fitness range in $\mathcal{A}$ within each benchmark shown in~\cref{tab:datasets}. For \gls{aav}, we therefore set $\mathcal{I} = [0.05, 0.5]$ and for \gls{gfp}, $\mathcal{I} = [0.05, 0.8]$. For all benchmarks, we used $|\mathcal{F}| = 20$ evenly spaced target values. The label perturbation scales were set to $q=0.0075$ for \gls{aav} and $q=0.01$ for \gls{gfp} to ensure dense coverage without artificial gaps between the target bins.
To generate the augmented dataset $\mathcal{A}_{\text{aug}}$, we used models trained with score dropout $p=0.1$ for the AAV benchmarks (Medium and Hard) and $p=0$ for the GFP benchmarks (Medium and Hard). All samples were generated with the full conditional distribution setting $w=0$. The flow model $v_\theta$ was subsequently retrained on $\mathcal{A}_{\text{aug}}$ using score dropout $p=0.1$ for regularization due to additional synthetic labels. The target fitness values were $\{0.52, 0.53, 1.05, 1.3\}$ and the \gls{cfg} parameter was set to $\{-0.1, 0.1, 0.0, 0.2\}$ for \gls{aav} Medium and Hard, and \gls{gfp} Medium and Hard, respectively. 

\section{Metrics}
\label{sec:metrics}
For evaluating the generated protein sequences, we follow the metrics described in previous works \cite{smoothgwg,gflow}: 
{median fitness}, {diversity}, and {novelty}. All scores are computed using the pretrained oracle $\Omega_{\phi}$, which was repurposed without any modification from~\cite{smoothgwg}, where it trained on the full ground-truth dataset $\mathcal{A}_{\text{GT}}$ with minimum and maximum fitness values denoted by $f_{\min}$ and $f_{\max}$.

\paragraph{Median Fitness.}  
For a set of generated sequences $S_i$ (where $i$ denotes one run), the normalized median fitness is computed as
\begin{equation}
    \mathrm{median}\left( \Big\{
        \frac{\Omega_{\phi}(x) - f_{\min}}{f_{\max} - f_{\min}}\  : \ x \in S_i
    \Big\} \right).
\end{equation}

\paragraph{Diversity.}  
Diversity is defined as the median pairwise distance between distinct generated sequences:
\begin{equation}
    \mathrm{median}\left( \Big\{ \mathrm{dist}(x, x') \ : \ x, x' \in S_i,\ x \neq x', \Big\} \right),
\end{equation}
where we use the Levenshtein distance, as we evaluate discrete protein sequences.

\paragraph{Novelty.}  
Novelty measures how different the generated sequences $S_i$ are from those in the dataset $\mathcal{A}$ used to train the model. It is defined as the median of the minimum distances from each generated sequence to any training sequence:
\begin{equation}
    \mathrm{median}\Bigl( \Big\{
        \min_{\hat{x} \in \mathcal{A}} \{ \mathrm{dist}(x, \hat{x}) \ :
    \ x \in S_i, \hat{x}\neq x \} \Big\}\Bigr).
\end{equation}

\section{Additional Results}
Finally, we benchmark the runtime and memory usage of \gls{method}. We compare against both variants of \gls{vlgpo} across all four benchmarks, as it is the most competitive baseline and provides published checkpoints and code. 
The results are summarized in \cref{table:runtime_memory}. While \gls{method} exhibits a 2- to 3-fold increase in peak memory consumption, its absolute memory footprint remains low. This represents a favorable trade-off for a substantial gain in efficiency, with inference times reduced by a factor of 10 to 85.

\begin{table*}[ht]
    \caption{Computational efficiency of predictor-guided vs. conditional sampling. Latency (s) represents the sampling time to generate 128 top-$k$ sequences and Memory (GB) represents peak GPU utilization.}
    \label{table:runtime_memory}
    \centering
    \resizebox{\linewidth}{!}{
    \begin{tabular}{lcccccccc}
        \toprule
        & \multicolumn{2}{c}{AAV Medium} & \multicolumn{2}{c}{AAV Hard} & \multicolumn{2}{c}{GFP Medium} & \multicolumn{2}{c}{GFP Hard} \\
        \cmidrule(lr){2-3} \cmidrule(lr){4-5} \cmidrule(lr){6-7} \cmidrule(lr){8-9}
        Method & Time (s) & Mem (GB) & Time (s) & Mem (GB)& Time (s) & Mem (GB)& Time (s) & Mem (GB)\\
        \midrule 
        VLGPO (unsmoothed) &  51.7 &0.25 & 24.9 & 0.25  & 78.3 & 1.17 & 10.6 &1.17  \\
         VLGPO (smoothed) & 24.9 & 0.25 & 25.5 & 0.25 &  10.9 & 1.17& 10.5& 1.17   \\
        CHASE & 0.63&  0.74&  0.53&  0.62&  1.00&   2.35&  0.91& 2.29\\
        \bottomrule
    \end{tabular}
    }
\end{table*}

\section{Additional Ablations}
\label{sec:additional_ablations}

\begin{figure*}[ht!]
    \centering
    \begin{subfigure}[b]{0.45\textwidth}
        \centering
        \includegraphics[width=\textwidth]{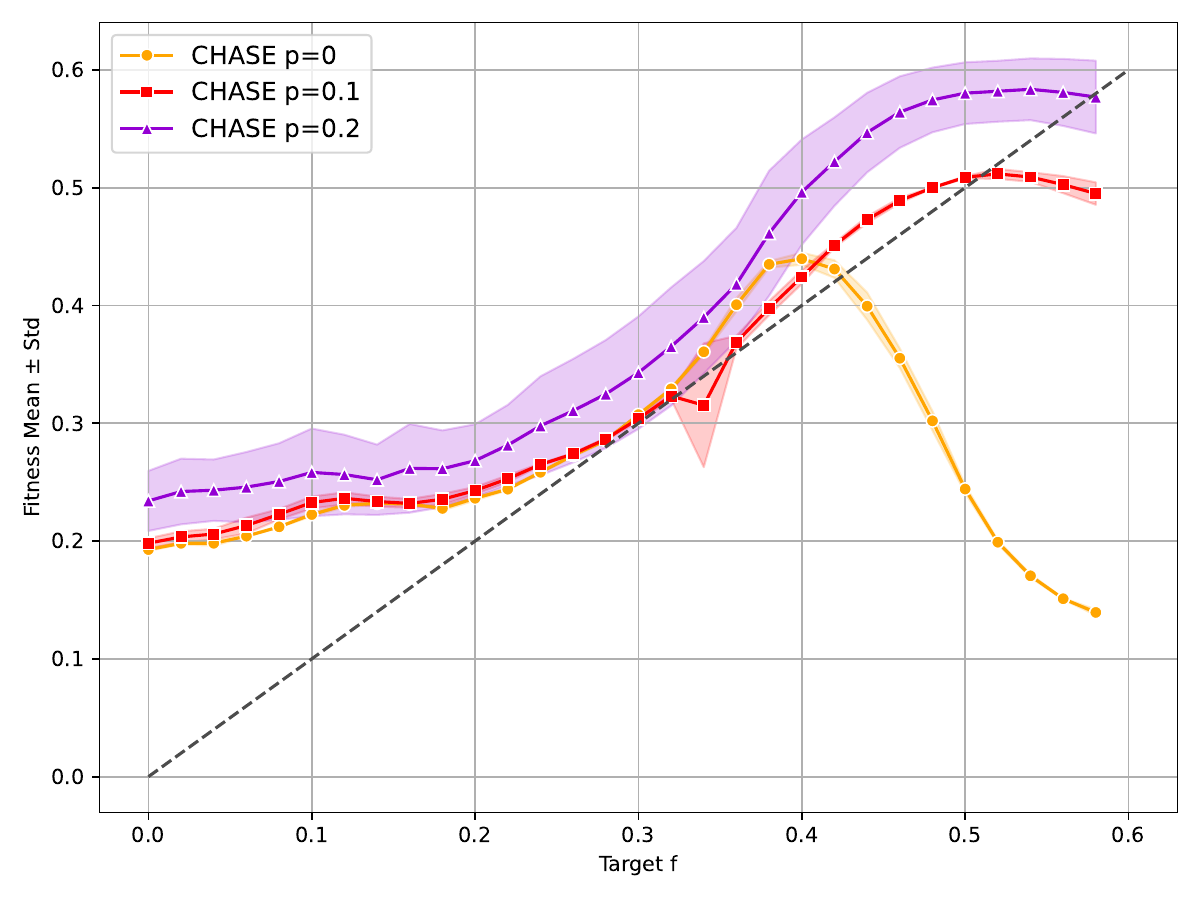}
        \caption{AAV Hard}
        \label{fig:aav_p_effect}
    \end{subfigure}
    \hspace{0.04\textwidth}
    \begin{subfigure}[b]{0.45\textwidth}
        \centering
        \includegraphics[width=\textwidth]{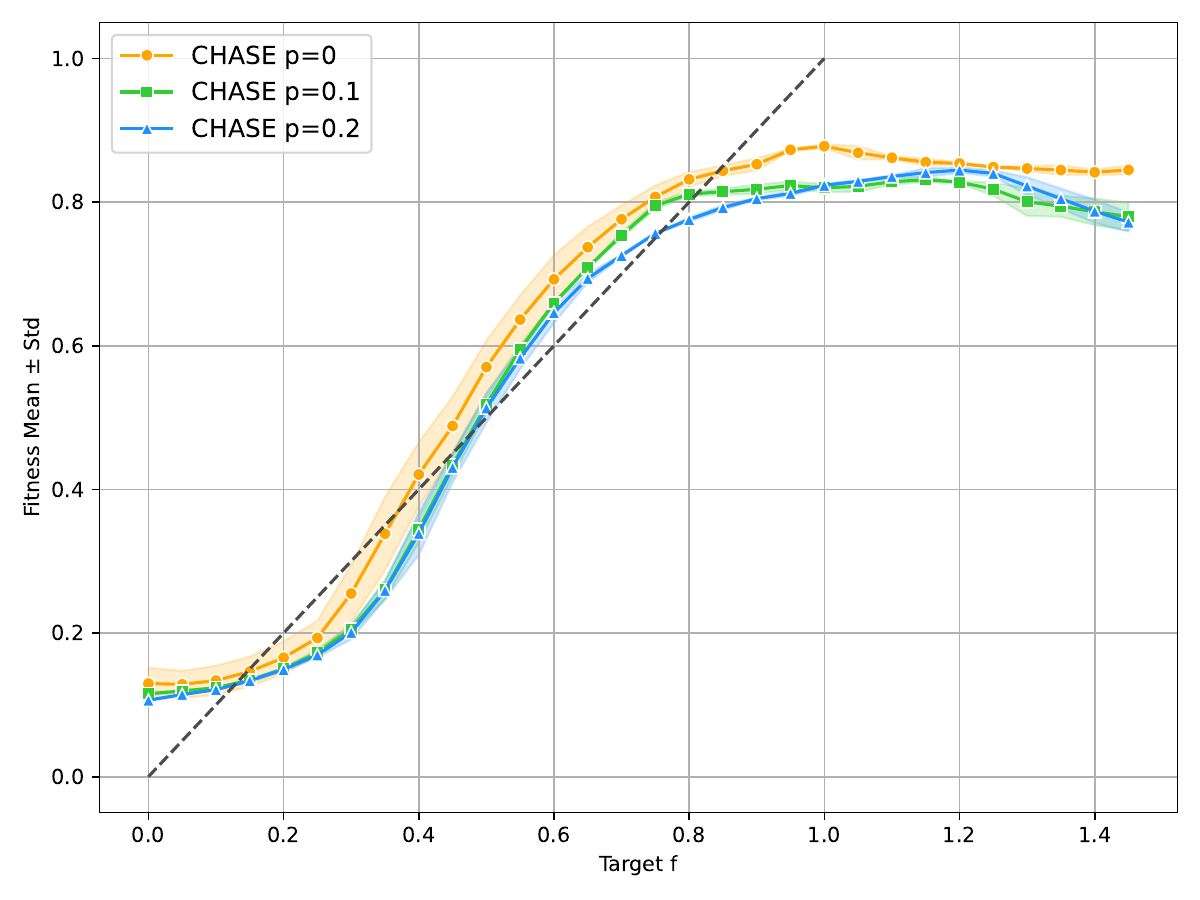}
        \caption{GFP Medium}
        \label{fig:gfp_p_effect}
    \end{subfigure}
    \caption{Effect of score dropout rate $p$ during training on model performance for AAV Hard and GFP Medium benchmarks.}
    \label{fig:p_ablation}
\end{figure*}

We additionally performed an ablation study on the impact of the dropout $p$ during flow matching training. This can be observed in~\cref{fig:p_ablation} for two selected benchmarks, contrasting the target fitness with the fitness of the generated sequences. While performance is generally stable across parameters, we observe that a minimum dropout can be beneficial for specific cases (e.g., \gls{aav} hard) to ensure more consistent behavior.

\end{document}